\documentclass[letterpaper, 10 pt, conference]{ieeeconf}  

\IEEEoverridecommandlockouts                              

\usepackage[utf8]{inputenc} 
\usepackage[T1]{fontenc}    
\usepackage{url}            
\usepackage{booktabs}       
\usepackage{amsfonts}       
\usepackage{xfrac}
\usepackage{microtype}      
\usepackage{xcolor}         
\usepackage{colortbl}       

\usepackage{listings}
\usepackage{soul}
\usepackage{lipsum}

\usepackage{graphicx, subfigure}
\usepackage{multirow, longtable, diagbox, setspace, rotating}
\usepackage{wrapfig, float, caption}

\usepackage{amsmath, amssymb, amsfonts} 
\usepackage{mathtools}
\usepackage[ruled, vlined]{algorithm2e}
\usepackage{algpseudocode}

\usepackage{tikz}
\usetikzlibrary{arrows.meta, positioning, shapes.geometric, calc, decorations.pathmorphing}

\title{\LARGE \bf
Keypose Exploration: Efficient Automatic Trajectory Labelling and Cross-Embodiment Policy Transfer
}

\author{
Yupu Lu$^{1}$, Hang Xu$^{1}$, Yizhou Chen$^{1}$ and Jia Pan$^{1}$%
\thanks{$^{1}$School of Computing and Data Science, The University of Hong Kong, HKSAR. 
        {\tt\small Correspondence: \{luyp16,panj\}@connect.hku.hk} }%
}

\begin{document}

\maketitle
\thispagestyle{empty}
\pagestyle{empty}

\begin{abstract}
Keypose-based manipulation decomposes tasks into critical waypoints to simplify policy learning for long-horizon tasks, but existing approaches rely on task-specific heuristics or manual annotation to extract keyposes from demonstrations.
We present an automatic trajectory labelling pipeline for grasp-related tasks. This pipeline combines vision-language models (VLMs) for semantic event detection with classical trajectory analysis for precise temporal alignment, requiring VLM inference only on one single demo among repeating ones per task.
Using the labelled data, we train a keypose-guided Diffusion Policy (DP) that exploits keypose conditioning to intervene demonstration distributions. We explore the possibility to apply this property for cross-embodiment transfer: candidate keyposes are sampled and filtered via a reachability map, steering the policy toward kinematically feasible keyposes for the target robot.
As a preliminary feasibility study, experiments on two robomimic tasks show that the labelled data produces policies matching a standard DP baseline, and that reachability-filtered keypose conditioning may benefit zero-shot transfer on the multimodal insertion task when feasible candidates are available.
\end{abstract}

\section{INTRODUCTION}

Keypose-based manipulation has emerged as an effective paradigm for long-horizon robotic tasks~\cite{gervet2023act3d, xian2023chaineddiffuser, ma2024hierarchical, yu2024bikc, xu2025bikc+}.
By decomposing a task into a sequence of critical end-effector poses (keyposes) and connecting them with continuous trajectories, these methods reduce the complexity of policy learning and improve interpretability.

A prerequisite for all such approaches is a reliable set of keypose labels in the demonstration data.
However, existing labelling methods require human intervention, either through hand-crafted heuristics or direct annotation.
Hand-crafted heuristics, such as gripper state transitions, velocity thresholds, or height conditions~\cite{gao2024bikvil, yu2024bikc}, demand case-by-case coding for each new task.
Direct manual annotation, such as decomposing multi-view videos into clips by human annotators and processing them with a VLM~\cite{wu2026pragmatic}, is more flexible but prohibitively slow and computationally costly for large-scale datasets.
Beyond the labelling bottleneck, how to further exploit the features of keypose for downstream policy learning, such as cross-embodiment transfer under multimodal demonstration distributions, remains an open question.

In this work we study \emph{grasp-related prehensile manipulation with parallel grippers} where keyposes correspond to interaction moments such as grasping, releasing, and object-object interaction. We treat the present results as a preliminary feasibility study rather than a fully general system.
We examine a two-stage automatic labelling pipeline that pairs a vision-language model (VLM) with classical trajectory analysis.
The VLM reasons about \emph{what} happens in a single demonstration video, identifying subtask semantics and interaction events, while a deterministic motion-segmentation module pinpoints \emph{when} each event occurs by aligning VLM-estimated frames to physical motion boundaries and gripper transitions.
This division requires only one VLM inference per task and rapidly propagates the inferred workflow to other same demonstrations without much additional human effort.

We also explore how to leverage the keypose conditioning mechanism for policy design. When demonstrations contain qualitatively different strategies, we introduce a reachability filter~\cite{lu2026richmap} that retains only kinematically feasible candidates for the target robot, enabling zero-shot cross-embodiment transfer without retraining.

Our contributions are as follows.

\textbf{Automatic trajectory labelling pipeline.}
We design a raw pipeline that leverages VLM comprehension for task-level semantic understanding and classical trajectory analysis for precise temporal alignment, producing keypose labels from demonstration data with minimal human intervention.

\textbf{Keypose-guided policy with reachability filtering.}
As a prototype policy, we integrate a reachability filter~\cite{lu2026richmap} into a keypose-guided Diffusion Policy to select kinematically feasible keyposes for the target robot, connecting high-level mode selection with low-level control.

\textbf{Preliminary experimental exploration.}
We conduct two robomimic experiments~\cite{robomimic2021} on Can and Square tasks as a scoping feasibility study for future exploration: validating that automatically labelled keyposes produce a policy matching the DP baseline, and investigating whether reachability-filtered keypose conditioning benefits zero-shot cross-embodiment transfer.

\section{RELATED WORKS}


\subsection{Keypose in Demonstration Trajectory}

An analysis of how humans approach multi-stage tasks indicates that people typically anticipate keyposes (i.e., upcoming target poses) and regard them as sub-goals. 
These keyposes serve to guide immediate actions within a shorter time frame~\cite{nakahashi2016modeling, binder2023humans}, thereby enabling the effective decomposition of complex tasks into several more manageable sub-stages. 
Similar ideas have been investigated in the field of robotic manipulation~\cite{ma2024hierarchical, yu2024bikc, xu2025bikc+}. 
By forecasting the next keypose and linking consecutive keyposes through continuous trajectories~\cite{gervet2023act3d, xian2023chaineddiffuser}, these approaches have achieved enhanced results in long-horizon tasks.

The accuracy and rationality of keypose identification are crucial for such keypose-based manipulation approaches. 
Nevertheless, most existing studies rely on heuristic rules to extract keyposes, summarized in previous research~\cite{gervet2023act3d, xian2023chaineddiffuser, gao2024bikvil, ma2024hierarchical, yu2024bikc, xu2025bikc+}. They include gripper open/close transitions, velocity thresholds, and relative distance or height conditions. 
These rules are hard-coded for each specific task and lack flexibility. 
An alternative is direct manual annotation~\cite{wu2026pragmatic}, where human annotators decompose demonstration videos into clips and mark interaction events; this approach is more flexible but prohibitively slow for large-scale datasets.
In comparison, our work aims to leverage a VLM to automatically identify keyposes from demonstration data, trying to use minimal task-specific definition while keeping the per-event keypose rules out of the labelling logic.

\subsection{Generative Imitation Learning}

Imitation learning (IL) serves as a primary approach to solving robotic manipulation tasks, enabling robots to directly acquire capabilities from expert demonstration data. 
The typical process comprises two core stages: (1) collecting demonstration data through teleoperation tools~\cite{zhao2023learning, chi2024universal, hirao2023body}; and (2) utilizing sophisticated parameterized models to approximate the state-action relationships reflected in the demonstration data~\cite{wang2024novel, xu2024leto}. 
Although IL is characterized by its simplicity and efficiency, it is susceptible to error accumulation~\cite{zhao2023learning} and encounters difficulties arising from the distributional multimodality of demonstration data~\cite{chi2023diffusionpolicy}.

As research advances, the integration of generative modeling into IL~\cite{zhao2023learning, chi2023diffusionpolicy} has yielded remarkable achievements in addressing these challenges. 
These approaches effectively mitigate cumulative errors by predicting action sequences to reducing the number of inference steps, and tackle the multimodal problem by leveraging generative models to approximate complex data distributions.
For instance, ACT~\cite{zhao2023learning} models the robot policy as a conditional variational autoencoder (cVAE), extracting core behavioral patterns from demonstrations. 
Diffusion Policy (DP)~\cite{chi2023diffusionpolicy} captures multimodal action distributions by representing the policy as a diffusion model that gradually refines action sequences from random noise, though the iterative denoising process leads to substantial inference latency.
Recently, several studies~\cite{lu2024manicm, prasad2024consistency, yu2024bikc, xu2025bikc+} have proposed Consistency Models (CMs)~\cite{song2023consistency, song2024improved} as a single-step alternative to diffusion models for manipulation policies, preserving the ability to capture distributional multimodality while offering much higher inference efficiency.

In this work, we adopt DP as the trajectory generation backbone for its strong capacity in capturing multimodal action distributions, and employ a CM as the keypose predictor for its efficient single-step inference, which enables rapid sampling of multiple keypose candidates during the propose-filter-generate pipeline.

\section{METHODOLOGY}
\label{sec:methodology}

We focus on \emph{grasp-related tasks} whose keyposes correspond to critical interaction moments, and explore two challenges:
(1)~automatic extraction of keyposes from demonstration data without manual annotation, and
(2)~expanding the utility of keyposes by integrating reachability analysis for better performance.
The pipeline consists of three components: VL-guided subtask analysis (Sec.~III-A), automatic trajectory-based correction (Sec.~III-B), and keypose-conditioned policy learning (Sec.~III-C,~III-D).

\begin{figure*}[t]
\vspace{0.15cm}
    \centering
    \includegraphics[width=\textwidth]{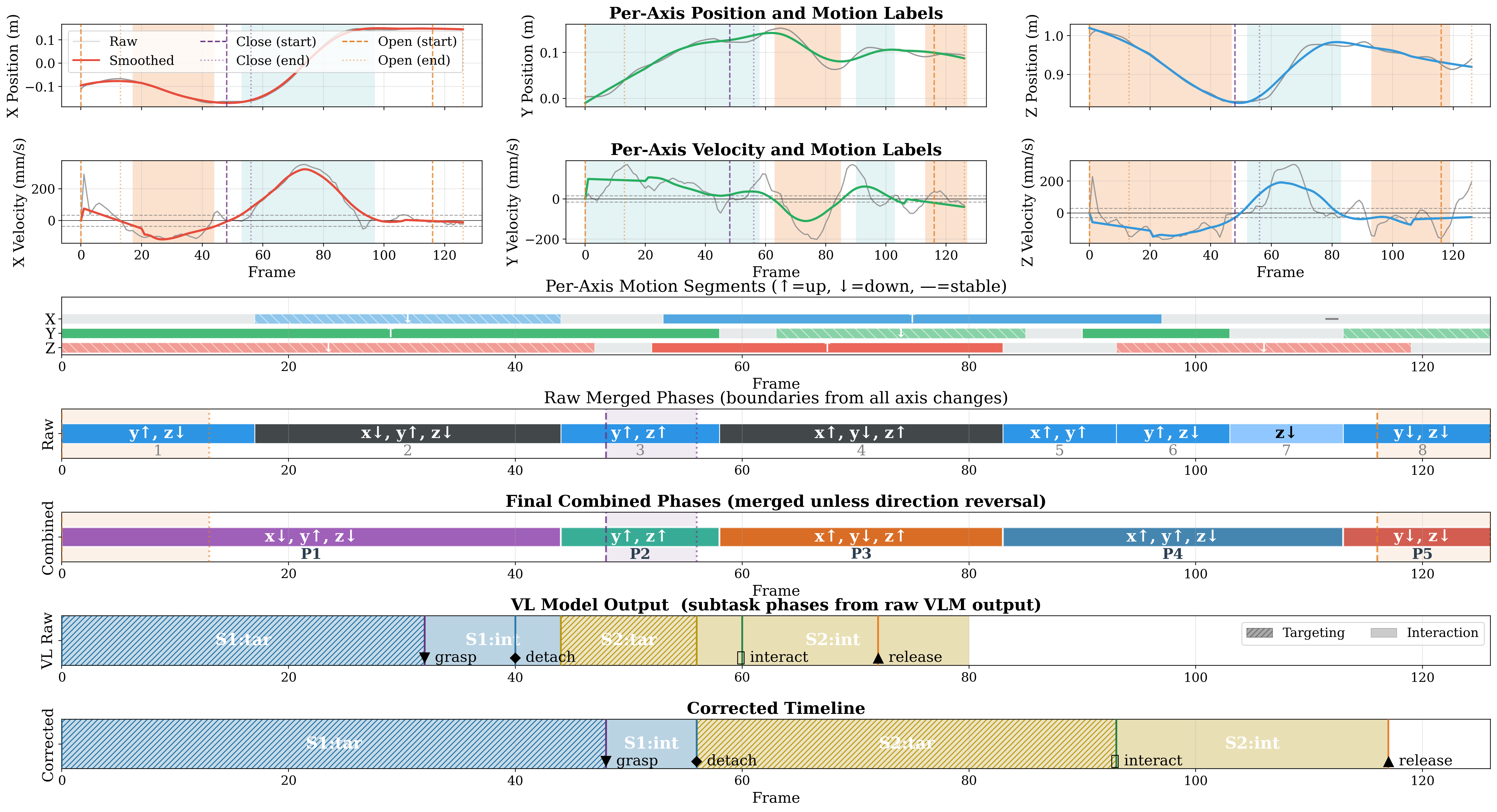}
    \caption{%
        Automatic labelling pipeline illustrated on a square peg-insertion episode.
        \textbf{Row~6 (Sec.~III-A):} VLM subtask analysis output.
        Each subtask S$i$ is decomposed into
        targeting~(\textit{hatched}), interaction~(\textit{solid}), and result~(\textit{dotted})
        phases, with event primitives (\texttt{grasp}~$\blacktriangledown$,
        \texttt{detach}~$\blacklozenge$, \texttt{interact}~$\square$,
        \texttt{release}~$\blacktriangle$, \texttt{land}~$\bullet$) placed at their VLM-estimated frames.
        \textbf{Rows~1--5 (Sec.~III-B):} Trajectory-based motion segmentation.
        Rows~1--3 show per-axis analysis (Stage~1): smoothed position and velocity
        curves are labelled per frame ($\uparrow$/$\downarrow$/stable) and independently segmented
        into directional intervals for each Cartesian axis.
        Rows~4--5 show cross-axis combination (Stage~2): all axis boundaries are first merged
        into a unified bar (Row~4), then adjacent phases are collapsed unless any axis reverses
        direction, yielding the final motion phases P1--P$n$ with gripper intervals overlaid (Row~5).
        \textbf{Row~7 (Sec.~III-B):} Corrected timeline after applying semantic-motion correspondence rules:
        event frames from Row~6 are snapped to gripper transitions and
        motion-phase boundaries identified in Rows~1--5.
    }
    \label{fig:trajectory_seg}
\end{figure*}

\subsection{Vision-Language Guided Subtask Analysis}

\subsubsection{Subtask Design for Grasp-Related Tasks}

We design a structured prompt that guides a vision-language model (Qwen3-VL-235B~\cite{Qwen3-VL}) to decompose manipulation demonstrations into subtasks. The input is a single demonstration provided as a batch of uniformly sampled RGB frames (the third-person video stream of one demo), passed to the VLM together with the task context. Each subtask follows a three-phase structure tailored for grasping operations:

\textbf{(1) Targeting phase:} Robot approaches the target object or goal location. This phase contains no object interactions and ends when positioned for manipulation.

\textbf{(2) Interaction phase:} Robot actively performs manipulation primitives:
\begin{itemize}
    \item \texttt{grasp}: gripper closes to hold object
    \item \texttt{release}: gripper opens to let go
    \item \texttt{detach}: object separates from surface during lifting
    \item \texttt{interact}: gripper-held object makes physical contact with a target object or surface (e.g., insertion tip touches peg, tool tip touches hole); the gripper remains closed
\end{itemize}

\textbf{(3) Result phase} (optional): Passive consequences, specifically \texttt{land} when a released object settles onto a surface after being dropped.

Each primitive is represented as $[\text{edge}, \text{type}, t]$ where $\text{edge}=[A, B]$ indicates interacting entities. The VLM receives task context and outputs structured JSON:

\begin{small}
\begin{verbatim}
{"subtask1": {"targeting": {...},
  "interaction": {"start": 8, "end": 12,
    "connections": [
      [["robot","can"], "grasp", 8],
      [["can","table"], "detach", 10]]}},
 "subtask2": {"targeting": {...},
  "interaction": {..., "connections": [
      [["robot","can"], "release", 23]]},
  "result": {..., "connections": [
      [["can","tray"], "land", 24]]}}}
\end{verbatim}
\end{small}

This prompting strategy enables consistent event detection across pick-place, tool hanging, and insertion scenarios while providing frame-level timestamps.

\subsection{Automatic Trajectory-Based Event Correction}

VLMs often produce temporal misalignments due to video frame rate reduction and perceptual ambiguities. We develop a \emph{hierarchical trajectory analysis framework} that automatically corrects these errors by establishing correspondences between semantic events and physical motion patterns.
The framework constructs three complementary timelines from the raw trajectory data: (1)~a \emph{motion timeline} obtained by double-layer segmentation that captures both fast and slow end-effector movements, (2)~a \emph{gripper timeline} recording opening and closing intervals, and (3)~a set of \emph{semantic-motion correspondence rules} that snap VLM-estimated event frames to precise positions in the first two timelines (see Fig.~\ref{fig:trajectory_seg}).

\subsubsection{Double-Layer Motion Segmentation}

We employ a \emph{segment-then-combine} strategy with per-axis motion analysis (Fig.~\ref{fig:trajectory_seg}, Rows~1--5).
The segmentation is applied in two layers: the first layer captures fast motions, and the second layer re-examines the resulting stable phases to locate finer actions.

\textbf{Stage 1 - Per-Axis Segmentation:} For each Cartesian axis $k \in \{x, y, z\}$, we first apply Savitzky-Golay smoothing to reduce sensor noise while preserving motion structure:
\begin{equation}
    \tilde{p}_t^k = \text{SavGol}(p_t^k; w, d), \quad v_t^k = \frac{\tilde{p}_{t+1}^k - \tilde{p}_t^k}{\Delta t}
\end{equation}
where $w$ is the window size (task-dependent, typically 41--81 frames) and $d$ is the polynomial order.

Then we apply per-axis velocity thresholds $\theta^k$ calibrated to task-specific motion ranges. Each frame receives a motion label:
\begin{equation}
    \ell_t^k = \begin{cases}
        +1 \text{ (up)} & \text{if } v_t^k > \theta^k \\
        -1 \text{ (down)} & \text{if } v_t^k < -\theta^k \\
        0 \text{ (stable)} & \text{otherwise}
    \end{cases}.
\end{equation}
\begin{figure}[H]
    \centering
    \includegraphics[width=\columnwidth]{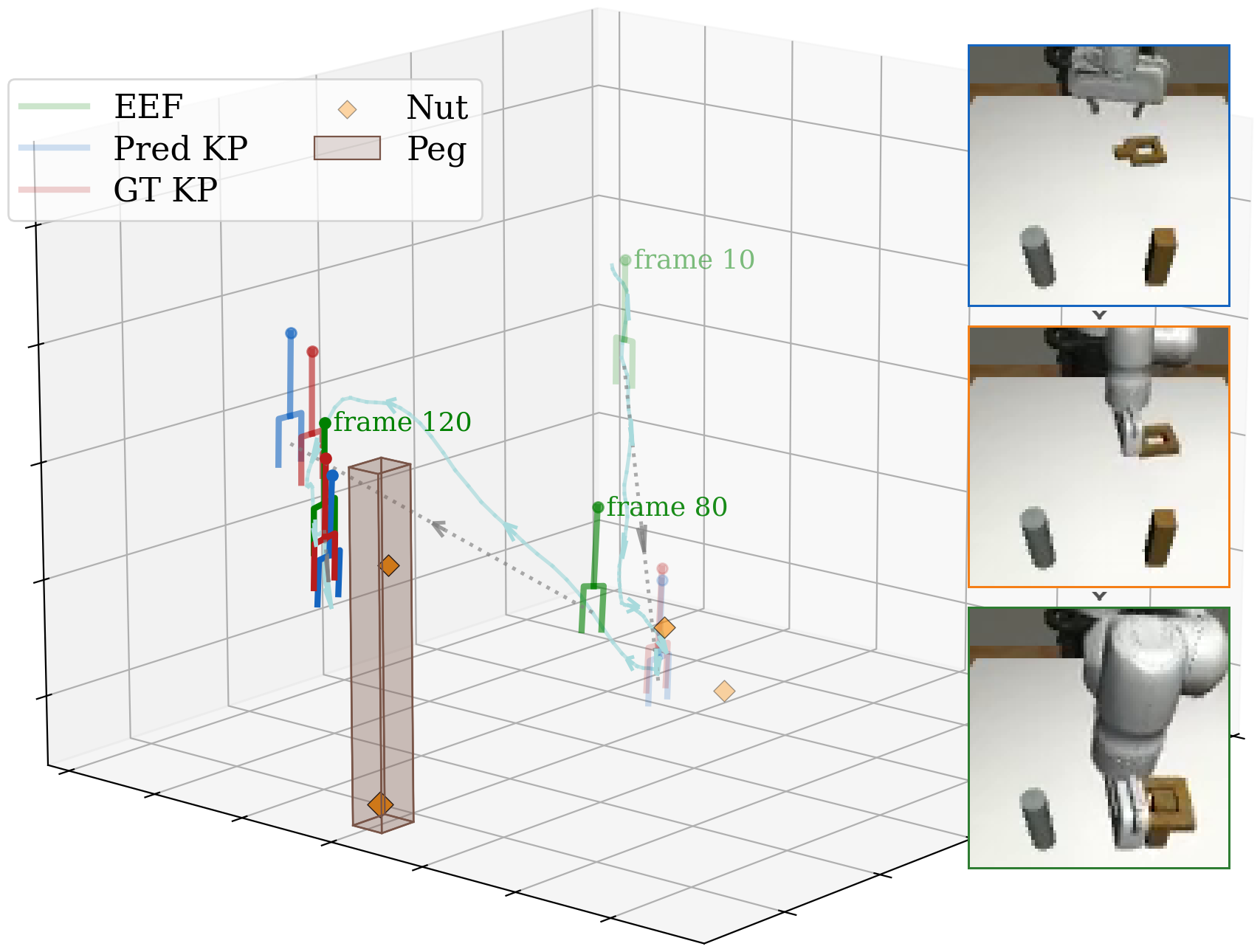}
    \caption{One square peg-insertion demo echos Fig.~\ref{fig:trajectory_seg} .
    Keypose predictor output along the EEF trajectory (teal).
    At each key frame the predicted keypose (blue gripper) is shown against the
    ground-truth (red gripper); opacity increases over time.
    Orange diamonds: nut; brown cuboid: peg.
    Three simulation frames (top to bottom, right panel) show the progression:
    targeting, grasping, and interacting.}
    \label{fig:kp_vis}
\end{figure}
The window $w$ and the per-axis thresholds $\theta^k$ are the only per-task quantities (their values are reported in Sec.~\ref{sec:experiments}); all remaining parameters below are global constants shared across tasks.

Finally, segment each axis independently into intervals of consistent motion direction.

\textbf{Stage 2 - Cross-Axis Combination:} We merge per-axis segments into unified motion phases using direction-aware logic:
\begin{itemize}
    \item \emph{Merge} adjacent phases unless any axis \emph{reverses} direction ($+1 \leftrightarrow -1$). Starting or stopping motion ($0 \leftrightarrow \pm 1$) does not break continuity.
    \item \emph{Discard} segments shorter than a minimum length (20 frames, 1s) and drop phase boundaries spaced closer than half that length, so that brief stable intervals caused by contact forces do not fragment a continuous action.
\end{itemize}

\textbf{Optional Layer --- Subtle Motion Detection:} Within the stable phases produced by Stages~1--2, we repeat the segmentation logic at a finer scale: cumulative displacement over a sliding window is analysed, and if it exceeds a threshold fraction of the working range, the phase is reclassified as slow-moving with appropriate direction labels. This double-layer design ensures that both fast transits and slow deliberate motions (e.g., careful approach before insertion) are captured.

\textbf{Gripper timeline} is segmented independently by detecting opening/closing transitions from gripper commands. Together, the motion and gripper timelines yield:
\begin{align}
    \mathcal{T}_{\text{move}} &= \{\text{Interval}(t_s, t_e, \text{signature})\} \\
    \mathcal{T}_{\text{grip}} &= \{\text{Interval}(t_s, t_e, \{\text{open}, \text{close}\})\}
\end{align}

\subsubsection{Semantic-Motion Correspondence Rules}

With motion and gripper timelines established, we define deterministic rules that map VL-detected semantic events to precise trajectory frames. The key insight is that each event type has a characteristic physical signature:

\begin{itemize}
    \item \textbf{\texttt{grasp}}: First frame of gripper \emph{closing} interval.
    
    \item \textbf{\texttt{release}}: First frame of gripper \emph{opening} interval.
    
    \item \textbf{\texttt{detach}}: Object lifts after grasping. Corrected to:
    \begin{equation}
        t^*_{\text{detach}} = \max(t_{\text{close\_end}}, t_{\text{move\_start}})
    \end{equation}
    where $t_{\text{close\_end}}$ is the end of closing and $t_{\text{move\_start}}$ is the first moving frame after grasp.
    
    \item \textbf{\texttt{interact}}: Gripper-held object makes physical contact with the target
    (insertion, hanging, placement while held).
    The gripper \emph{remains closed} throughout.
    Corrected to the \emph{first frame of the last moving phase before release}.
    
    \item \textbf{\texttt{land}}: Released object passively settles onto a surface.
    The gripper is \emph{already open}; this event is a consequence of \texttt{release},
    not an active robot action.
    Corrected to the end of the opening interval or subtask boundary.
\end{itemize}

These rules leverage the motion signatures for context-aware matching. For instance, \texttt{interact} in an insertion task should align with a phase containing downward z-motion ($z\!\downarrow$), while tool hanging may involve horizontal approach patterns.

After the correction, the events yield keypose frames $\{t_1^*, \ldots, t_K^*\}$ aligned with physical interaction moments.
At each keypose frame we record the end-effector pose $\mathbf{k}_i = [\mathbf{p}_i, \mathbf{q}_i, g_i]$, where $\mathbf{p}_i\!\in\!\mathbb{R}^3$ is the gripper-site position in the world frame, $\mathbf{q}_i$ the end-effector orientation, and $g_i\!\in\!\mathbb{R}^2$ the two finger joint positions.
The orientation is stored as a quaternion but fed to the networks as a 6D continuous rotation~\cite{zhou2019continuity} to avoid the quaternion double cover, giving an 11D target vector $[\mathbf{p}(3), \mathbf{r}_{6\text{D}}(6), g(2)]$.

Targets are normalized per dimension to $[-1, 1]$ (min--max, with a small-variance guard for near-constant channels).
The joint configuration at each keypose frame is also recorded but is not used as a network input in the present experiments.

\subsection{Keypose Prediction Networks}

Building on the BiKC framework~\cite{xu2025bikc+} that decomposes trajectory-level imitation learning into keypose prediction and trajectory generation, we train a keypose predictor on the automatically extracted labels.
Given extracted dataset $\mathcal{D} = \{(\mathbf{o}_j, \mathbf{k}_j)\}_{j=1}^{N}$, where $\mathbf{o}_j$ is an observation vector characterizing the current environment state and $\mathbf{k}_j$ is the target keypose, the network learns to predict the next keypose from the current observation.

The observation $\mathbf{o}$ is first processed by an encoder network, producing an observation embedding $\mathbf{f}_{\text{obs}}$.
We adopt a Consistency Model (CM)~\cite{song2023consistency, song2024improved} as the keypose predictor: a conditional denoiser $f_\theta(\mathbf{k}_t, t, \mathbf{f}_{\text{obs}})$ is trained via EMA-based consistency distillation~\cite{song2024improved} to map any noisy keypose $\mathbf{k}_t$ directly to the clean target $\mathbf{k}_0$, with the observation embedding conditioning the denoiser through feature modulation.
The consistency distillation loss is:
\begin{equation}
    \mathcal{L}_{\text{kp}} = \mathbb{E}_{\mathbf{k}_0,\, t} \big[\, d\big(f_\theta(\mathbf{k}_t, t, \mathbf{f}_{\text{obs}}),\, f_{\theta^-}(\mathbf{k}_{t'}, t', \mathbf{f}_{\text{obs}})\big) \big]
\end{equation}
where $t' < t$ is a neighbouring noise level, $\theta^-$ is the EMA target network, and $d(\cdot)$ is a pseudo-Huber metric.
A lightweight auxiliary head predicts the current subtask stage from $\mathbf{f}_{\text{obs}}$ via a cross-entropy term $\mathcal{L}_{\text{stage}}$, and the predictor is trained on $\mathcal{L}_{\text{kp}} + \lambda\,\mathcal{L}_{\text{stage}}$; this stage signal stabilises which keypose the predictor should target at inference.
At inference, the CM's single-step generation enables rapid sampling of $N_{\text{cand}}$ candidate keyposes, which is essential for the downstream propose-filter-generate pipeline.

\subsection{Keypose-Guided Diffusion Policy with Reachability Constraints}

We explore the keypose-guided conditional diffusion model~\cite{chi2023diffusionpolicy} where keyposes act as explicit conditions, decoupling high-level mode selection (keypose) from low-level control (trajectory).

\subsubsection{Policy Architecture and Training}

Let $\boldsymbol{\tau} = (\mathbf{a}_1, \ldots, \mathbf{a}_T)$ denote the action trajectory and $\mathbf{o}$ the current observation.
The trajectory policy $p(\boldsymbol{\tau} | \mathbf{o}, \mathbf{k})$ is conditioned on both the observation and a selected keypose $\mathbf{k}$, coupling high-level goal specification with low-level action generation.

The keypose $\mathbf{k}$ is injected by concatenating it feature-wise with $\mathbf{o}$ at each observation time step to form a combined condition $\mathbf{c} = [\mathbf{o};\, \mathbf{k}]$, which serves as the cross-attention memory in a Transformer-based denoising backbone.
The policy is trained with the standard noise prediction objective:
\begin{equation}
    \mathcal{L}_{\text{dp}} = \mathbb{E}_{t, \mathbf{o}, \mathbf{k}, \boldsymbol{\tau}_0, \boldsymbol{\epsilon}} \left[ \| \boldsymbol{\epsilon} - \boldsymbol{\epsilon}_\theta(\boldsymbol{\tau}_t, t, \mathbf{c}) \|^2 \right]
\end{equation}
where $\boldsymbol{\tau}_t$ is the noisy trajectory at diffusion step $t$.

\subsubsection{Reachability Filtering and Transfer}

To transfer to a new robot embodiment, we query a precomputed reachability map~\cite{lu2026richmap} that returns, for a given end-effector pose, whether it is kinematically reachable by the target robot. The map yields a binary suitability indicator $\mathcal{S}(\mathbf{k}; \text{robot})\!\in\!\{0,1\}$, which reweights the source keypose distribution:
\begin{equation}
    p_{\text{target}}(\mathbf{k} | \mathbf{o}) \propto \mathcal{S}(\mathbf{k}; \text{target}) \cdot p_{\text{source}}(\mathbf{k} | \mathbf{o})
\end{equation}
where $p_{\text{source}}(\mathbf{k} | \mathbf{o})$ is the keypose predictor distribution learned from demonstrations.
For the source robot, all demonstrated keyposes are reachable by definition; for a new embodiment, $\mathcal{S}$ restricts the policy to keyposes feasible for that robot before trajectory generation.


In practice, the decoupling of keypose selection and trajectory generation yields a propose--filter--generate transfer procedure.
First, $N_{\text{cand}}\!=\!24$ candidate keyposes are \textbf{proposed} by sampling the CM predictor $p_{\text{source}}(\mathbf{k}|\mathbf{o})$, each carrying a manifold-confidence score (the negative single-step denoising residual).
The candidates are then \textbf{filtered}: they are clustered into up to three orientation modes, each mode is queried against the reachability map, and we select the mode with the highest composite value (number of reachable candidates $\times$ best confidence), keeping its highest-confidence reachable candidate as $\mathbf{k}^*$.
If no candidate is reachable, the procedure \emph{falls back} to the highest-confidence candidate overall, so the policy always receives a keypose.
Finally, the frozen diffusion policy \textbf{generates} the action trajectory conditioned on $\mathbf{k}^*$.

This pipeline is designed to test zero-shot transfer without retraining, since the reachability constraint is applied entirely in keypose space.
When many candidates are reachable the filter could help steer the mode, but when feasible candidates are scarce the fallback dominates and the filter reduces to plain keypose prediction (Sec.~\ref{sec:experiments}).

\section{EXPERIMENTS}
\label{sec:experiments}

\subsection{Setup}

Our preliminary feasibility study pursues two goals:
(1)~\emph{validate} that the automatically labelled keypose data produces a working policy, i.e., the keypose-guided Diffusion Policy (KP-DP) performs on par with a standard DP trained on the same demonstrations; and
(2)~\emph{investigate} whether the multimodal structure exposed by keypose conditioning can be exploited to improve zero-shot cross-embodiment transfer.

We evaluate on two tasks from the robomimic environment~\cite{robomimic2021}.
\textit{Can} is a pick-and-place task with fixed placing positions and a single natural grasp direction; it is effectively unimodal and serves as a sanity check.
\textit{Square} requires inserting a square nut onto a peg; the nut inherits three distinct insertion orientations, and human demonstrators follow qualitatively different strategies across episodes, making the demonstration distribution intrinsically multimodal.
This contrast lets us isolate the effect of keypose conditioning on multimodal data.

All models are trained exclusively on 200 Franka Panda demonstrations and evaluated under zero-shot transfer to two target robots, \textbf{Kinova3} and \textbf{UR5e}, which differ from Panda in kinematic structure, link lengths, and reachable workspace.

To isolate the multimodal and reachability effects of keypose conditioning rather than visual domain shift, all observation inputs are deliberately provided as low-dimensional state vectors (object states and end-effector poses, $23$D, in the world frame) rather than camera images. This also avoids rendering discrepancies across embodiments. The action space is the $10$D absolute end-effector pose (position, 6D rotation, gripper) in the world frame, and normalization statistics are fit on the Panda source data and shared across all robots.

Three conditions are compared.
\textbf{DP}~\cite{chi2023diffusionpolicy}: the standard Diffusion Policy serving as the baseline.
\textbf{KP}: our keypose-guided policy, trained on demonstrations using automatically extracted keyposes (Sec.~\ref{sec:methodology}), transferred without reachability filtering.
\textbf{KP+Reach}: KP augmented with the propose--filter--generate pipeline, which down-weights candidate keyposes that are kinematically infeasible for the target robot before conditioning the policy (parameters below).
The keypose predictor is a lightweight consistency model (0.37\,M parameters); the trajectory policy is a diffusion transformer (8.97\,M parameters).

\textbf{Implementation details.}
The labelling pipeline shares one global configuration across tasks---smoothing order $d\!=\!2$, minimum segment length $20$ frames ($1.0$\,s), gripper-gap fill $\leq\!5$ frames ($0.25$\,s), and the subtle-motion window of $100$ frames ($5$\,s) with a $10\%$-of-working-range trigger---and only the smoothing window $w$ and per-axis velocity thresholds $\theta^k$ are set per task (Table~\ref{tab:seg_params}).
For transfer, the propose--filter--generate stage samples $N_{\text{cand}}\!=\!24$ candidates, clusters them into at most three orientation modes (split when candidates differ by $>\!20^\circ$ or $>\!5$\,cm), and queries a RichMap~\cite{lu2026richmap} built at a $0.05$\,m grid resolution.

\begin{table}[h]
\centering
\caption{Per-task labelling parameters: the smoothing window $w$ and per-axis velocity thresholds $\theta^k$ are set per task from each task's velocity profile. The subtle-motion displacement threshold is a shared constant ($10\%$ of each axis' working range). Demo trajectories are recorded at $20$\,Hz.}
\label{tab:seg_params}
\resizebox{.9\columnwidth}{!}{
\begin{tabular}{l|c|ccc|c}
\toprule
\textbf{Task} & $w$ & $\theta^x$ & $\theta^y$ & $\theta^z$ & disp.\ thr. \\
\midrule
Can    & 61 & 0.030 & 0.070 & 0.040 & \multirow{2}{*}{$0.10\times$range} \\
Square & 41 & 0.035 & 0.015 & 0.030 & \\
\bottomrule
\end{tabular}
}
\end{table}

\subsection{Results and Analysis}
All results are mean$\pm$std over 9~checkpoints (3~training seeds~$\times$~3~checkpoint epochs), each evaluated on 50~episodes, shown in Table~\ref{tab:transfer_simple}.

\begin{table}[h]
\centering
\vspace{0.2cm}
\caption{Success rate for zero-shot transfer from Panda to Kinova3 and UR5e on Can and Square.}
\label{tab:transfer_simple}
\resizebox{0.45\textwidth}{!}{%
\begin{tabular}{l|c!{\color{gray!60}\vrule}cc}
\toprule
\multicolumn{4}{c}{\textbf{Can}} \\
\cmidrule(lr){1-4}
\textbf{Method} & Panda & Kinova3 & UR5e \\
\midrule
DP        & 0.987$\pm$0.014          & \textbf{0.911$\pm$0.055}  & \textbf{0.96$\pm$0.022} \\
KP        & 1.000                    & 0.847$\pm$0.034           & 0.891$\pm$0.028 \\
KP+Reach  & 0.989$\pm$0.014          & 0.843$\pm$0.025           & 0.913$\pm$0.038 \\
\midrule\midrule
\multicolumn{4}{c}{\textbf{Square}} \\
\cmidrule(lr){1-4}
\textbf{Method} & Panda & Kinova3 & UR5e \\
\midrule
DP        & 0.862$\pm$0.036           & 0.220$\pm$0.045           & 0.738$\pm$0.103 \\
KP        & 0.842$\pm$0.052           & 0.187$\pm$0.037           & 0.758$\pm$0.055 \\
KP+Reach  & \textbf{0.873$\pm$0.035}  & \textbf{0.242$\pm$0.042}  & \textbf{0.829$\pm$0.044} \\
\bottomrule
\end{tabular}%
}
\end{table}

\begin{figure*}[t]
\vspace{0.15cm}
    \centering
    \includegraphics[width=\textwidth]{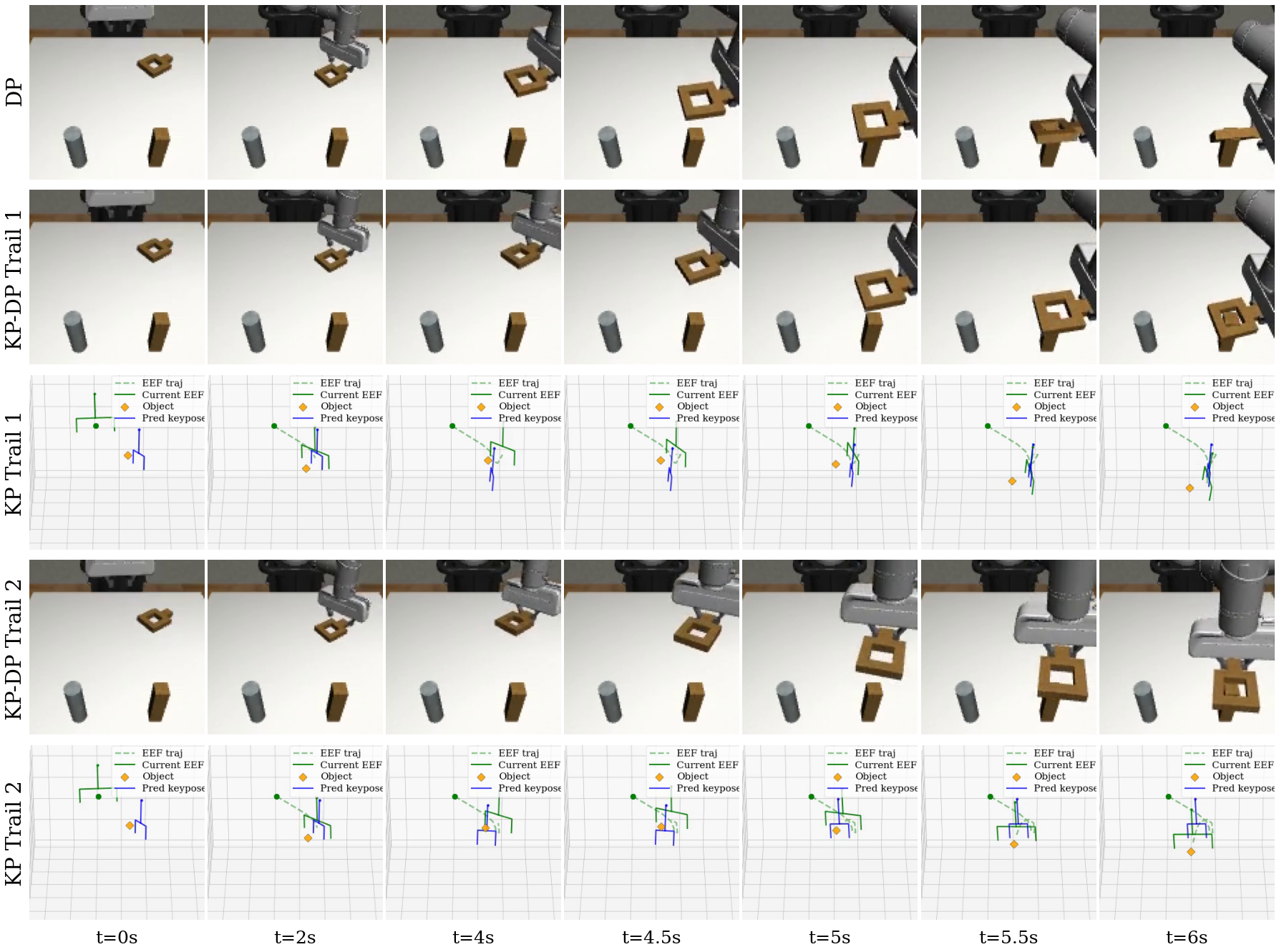}
    \caption{Rollout comparison for Square on UR5e across seven time stamps ($t$\,=\,0--6\,s).
    \textbf{Row~1 (DP):} It performs a direct insertion, but fails in this trial due to misaligned contact from $t=5$\,s.
    \textbf{Rows~2\,\&\,4 (KP-DP):} Two independent trials under KP+Reach that select different insertion orientations, illustrating mode switching.
    \textbf{Rows~3\,\&\,5 (KP scene):} Corresponding 3-D keypose trajectories; the blue gripper shows the predicted next keypose guiding the policy at each step.
    We also observe the more deliberate approach near the keypose (both grasping and insertion) in Rows~2\,\&\,4 versus the faster, direct motion of DP in Row~1.}
    \label{fig:comparison_frames}
\end{figure*}

\subsubsection{Labelling Validation}

We first verify the quality of the automatically labelled data by evaluating all three conditions on the Panda source robot.
KP and KP+Reach achieve 100\% and 98.9\% on Can, and 84.2\% and 87.3\% on Square, all closely matching the DP baseline (98.7\% and 86.2\%, respectively), confirming that the pipeline produces correct keypose annotations and that the KP-DP framework trains successfully on them.

The Square results further demonstrate that the keypose carries semantically meaningful information, not merely a redundant label.
KP+Reach is comparable to the DP baseline on Square ($87.3\%$ vs.\ $86.2\%$ on Panda; $82.9\%$ vs.\ $73.8\%$ on UR5e).
Since the reachability filter operates exclusively in keypose space (Sec.~III-D), any improvement in task success must propagate through the chain: keypose selection $\rightarrow$ policy conditioning $\rightarrow$ action generation.
The observed gain is a positive sign that the keypose functions as an active conditioning signal can shape the policy's output distribution.

On the Can task, the unimodal demonstration distribution and near-perfect DP baseline ($98.7\%$) leave little room for keypose conditioning to contribute: the policy is already well-calibrated from observations alone, and the additional conditioning signal introduces slight disruptions rather than improvements.
Consequently, KP and KP+Reach trail DP on Can across all robots, though the margin remains small.

\subsubsection{Multimodal Feature for Policy Transfer}

The Square task is where keypose conditioning demonstrates its utility.
The demonstration dataset contains three qualitatively different insertion orientations, making the action distribution intrinsically multimodal.
DP transferred to UR5e achieves $73.8\%$, a notable drop from the Panda baseline of $86.2\%$ with high variance ($\pm9.3\%$), indicating inconsistent failures across checkpoints and insertion modes.

We attribute the improvement brought by keypose conditioning to two mechanisms observed in our rollouts (Fig.~\ref{fig:comparison_frames}), described below.

\textbf{Mode selection via keypose conditioning.}
Conditioning the diffusion policy on a predicted keypose leads each rollout to one specific insertion orientation rather than averaging across modes.
The reachability filter works to ensure the selected mode is kinematically feasible for the target robot, steering the policy away from orientations that are geometrically hard to achieve or even infeasible.
This effect is visible in Fig.~\ref{fig:comparison_frames} Trial~1 (Rows~2--3) and Trial~2 (Rows~4--5) adopt different insertion directions yet both succeed, confirming that the keypose predictor can sample diverse modes and that the reachability filter selects appropriate ones for the target embodiment.

\textbf{Conservative behaviour near keyposes.}
The keypose-guided policy exhibits more deliberate motion as the end-effector approaches the predicted keypose, in contrast to DP's faster, more direct execution.
As illustrated in Fig.~\ref{fig:comparison_frames}, comparing Row~1 (DP) with Row~2, 4 (KP-DP): at $t{=}4$\,s, DP has already transported the nut to the peg, whereas KP-DP has just completed grasping; between $t{=}5{-}5.5$\,s, DP is already performing the insertion action while KP-DP is still in the targeting phase, approaching the peg more cautiously.
The keypose acts as a local attractor that focuses the policy's output distribution around the contact region, reducing positional error where precision matters most.
For peg insertion, this caution is beneficial: DP sometimes executes the demonstrated motion too aggressively, leading to misaligned contact and insertion failure.

Combining both mechanisms, KP alone stabilises transfer to $75.8\%$ (roughly halving the variance of DP), and KP+Reach further improves to $82.9\%$, within $3.3\%$ of the Panda source-robot baseline.

It is important to note that the keypose acts as a \emph{conditioning signal}, not a hard constraint that fully controls the policy, and its benefit is bounded by how many candidate keyposes are actually feasible for the target robot.
The Kinova3 results make this explicit. Measuring candidate feasibility on held-out states (sampling $24$ candidates per state and querying the reachability map), UR5e admits a reachable candidate at essentially every step ($\sim\!99\%$ acceptance on Square), whereas for Kinova3 almost no sampled candidate is reachable ($\sim\!0.1\%$ acceptance; the procedure falls back to the highest-confidence candidate at $>\!99\%$ of steps). Kinova3's reachable workspace and wrist limits are poorly matched to Square's insertion orientations, which arrive predominantly from the left/right sides.
Consequently KP+Reach on Kinova3 reduces almost everywhere to plain KP, and the change relative to DP ($22.0\%\!\rightarrow\!24.2\%$, $+2.2\%$) is within the run-to-run variation ($\pm4.2\%$/$\pm4.5\%$) and should not be read as a significant reachability effect.
The honest takeaway is that the filter helps when feasible candidates exist (UR5e), and that overcoming a large kinematic gap such as Kinova3's would require treating the keypose as a stronger constraint than the current soft conditioning.

\section{CONCLUSION}

We presented an automatic trajectory labelling pipeline for grasp-related manipulation tasks that combines vision-language semantic understanding with deterministic trajectory analysis, producing keypose annotations without manual annotation or per-event hand-coded keypose rules.
The resulting labels train a prototype keypose-guided Diffusion Policy that matches the performance of a standard baseline, validating the labelling quality.
On a multimodal insertion task, we showed that keypose conditioning exposes and disambiguates action modes, and that filtering candidate keyposes through a reachability map could improve zero-shot cross-embodiment transfer.

\subsection{Limitations and Future Work}

\emph{Scope of the keypose abstraction.} The more fundamental limitation is the keypose representation itself. Reducing a task to a sparse set of critical end-effector poses fits grasp-centric manipulation well, where grasping, releasing, and object--object contact define natural sub-goals, but it is somewhat hard to transfer to many manipulation tasks where no discrete interaction pose captures the essential behaviour, e.g.\ pushing, pivoting, wiping, pouring, or other continuous-contact and non-prehensile actions. Keypose labelling is therefore best viewed as a \emph{special case} of the broader problem of decomposing demonstrations into semantically meaningful \emph{subtasks}: a subtask formulation subsumes keyposes (an interaction subtask reduces to its boundary pose) while also describing actions that have no single defining pose. Generalising the present pipeline toward such subtask-level labelling, with keyposes retained only where they are the appropriate abstraction, is the main direction we are pursuing beyond this study.

\emph{Labelling assumptions.} Within the grasp-related setting, the pipeline still relies on a per-task smoothing window and per-axis velocity thresholds rather than being fully parameter-free. It runs the VLM on a single representative demonstration and propagates that workflow to the whole set, which assumes all demonstrations follow the same sequence of subtasks and execute it without mistakes. Teleoperated data with retries, hesitations, or reordered substeps would break this assumption, and the current rule-based semantic--motion correspondence does not yet reconcile a VLM-inferred workflow that conflicts with the observed motion signatures.

\emph{Policy and experiments.} 1) Our experiments are prototype-scale (two simulated tasks, low-dimensional states, two target robots), so the results indicate feasibility rather than broad generality. Both tasks are also relatively short-horizon, with only one or two interaction events each; the central benefit of decomposing a task into keyposes, structuring \emph{long-horizon}, multi-stage behaviour into a sequence of sub-goals, is therefore not fully exercised here. Evaluating on longer-horizon tasks would more clearly reveal the value of this division. 2) The keypose acts as a soft conditioning signal rather than a hard constraint, which limits its influence when the embodiment gap is large: on Kinova3 almost no sampled keypose is reachable, the filter falls back to plain prediction, and transfer remains poor. The unreachable keyposes do not mean the task cannot be finished completely, since there exists action error tolerance around the keypose to finish each subtask. 3) The keypose predictor capacity (0.37\,M parameters) is intentionally matched to the small demonstration set (200 episodes) to avoid overfitting, but this also constrains the diversity and precision of the predicted keyposes; scaling to richer datasets and stronger predictors and policy model would better reveal the potential of keypose conditioning.

\emph{Vision.} To isolate the multimodal and reachability effects we deliberately restricted observations to low-dimensional states, which sidesteps the visual domain shift that is central to real cross-embodiment transfer. It also prohibits the expression ability of policy. Integrating visual-input adaptation techniques~\cite{chen2024mirage, ji2025oxe} to enable vision-based policies while preserving transferability is an important next step.



\section*{ACKNOWLEDGMENT}
This work involves GitHub Copilot and Claude Code for both code developments and documentation preparation except bibliography. Careful line-by-line review and  adjustments were made for clarity and accuracy for the whole contents in this paper.

This work was supported by the Natural Science Foundation of China (62461160309), the NSFC-RGC Joint Research Scheme (N\_HKU705/24), Hong Kong RGC (GRF 17201025, GRF17200924). Yupu Lu was sponsored by the HKU Presidential PhD Scholarship.

\bibliographystyle{IEEEtran}
\bibliography{keypose_labelling}

\end{document}